\documentclass[letterpaper, 10 pt, journal, twoside]{IEEEtran}

\usepackage{amsmath,amsfonts}
\usepackage{algorithmic}
\usepackage{algorithm}
\usepackage{array}
\usepackage[caption=false,font=normalsize,labelfont=sf,textfont=sf]{subfig}
\usepackage{textcomp}
\usepackage{stfloats}
\usepackage{url}
\usepackage{verbatim}
\usepackage{graphicx}
\usepackage{cite}
\usepackage{multirow}
\usepackage{orcidlink}
\hypersetup{hypertex=true,
	colorlinks=true,
	linkcolor=blue,
	anchorcolor=blue,
	citecolor=blue}
    
\DeclareRobustCommand{\orcidicon}{
	\begin{tikzpicture}
		\draw[lime, fill=lime] (0,0)
		circle[radius=0.16]
		node[white]{{\fontfamily{qag}\selectfont \tiny \.{I}D}};
	\end{tikzpicture}
	\hspace{-2mm}
}
\foreach \x in {A, ..., Z}{%
	\expandafter\xdef\csname orcid\x\endcsname{\noexpand\href{https://orcid.org/\csname orcidauthor\x\endcsname}{\noexpand\orcidicon}}
}

\begin{document}

\title{
    DASP: Self-supervised Nighttime Monocular Depth Estimation with Domain Adaptation of Spatiotemporal Priors
}

\author{\IEEEauthorblockN{Yiheng Huang$^{\orcidlink{0009-0007-9527-6569}}$,
		Junhong Chen$^{\orcidlink{0000-0002-4874-9550}}$,
        Anqi Ning$^{\orcidlink{0009-0002-7456-2447}}$, 
        Zhanhong Liang$^{\orcidlink{0009-0007-5428-0115}}$, 
		Nick Michiels$^{\orcidlink{0000-0002-7047-5867}}$, \\ 
		Luc Claesen$^{\orcidlink{0000-0003-0405-6290}}$, 
        and Wenyin Liu$^{\orcidlink{0000-0002-6237-6607}}$}

        \thanks{
            Yiheng Huang, Zhanhong Liang, and Wenyin Liu are with the College of Computer Science and Technology, Guangdong University of Technology, Guangzhou 510006, China (e-mail: huangyiheng.gdut@gmail.com; cw252128385@gmail.com; liuwy@gdut.edu.cn). (Corresponding author: Junhong Chen and Wenyin Liu.) } 
        \thanks{
            Junhong Chen is with the College of Computer Science and Technology, Guangdong University of Technology, Guangzhou 510006, China, and also with the Digital Future Lab, Flanders Make, Hasselt University, 3590 Diepenbeek, Belgium (e-mail: CSChenjunhong@hotmail.com).
        }
        \thanks{
            Anqi Ning is with the College of Engineering, Shantou University, Shantou 515063, China (e-mail: ninganqi.stu@gmail.com).}  
        \thanks{
            Nick Michiels is with the Digital Future Lab, Flanders Make, Hasselt University, 3590 Diepenbeek, Belgium.}
        \thanks{
            Luc Claesen is with Hasselt University, 3530 Diepenbeek, Belgium.}
}



\maketitle

\begin{abstract}
Self-supervised monocular depth estimation has achieved notable success under daytime conditions. However, its performance deteriorates markedly at night due to low visibility and varying illumination, e.g., insufficient light causes textureless areas, and moving objects bring blurry regions. To this end, we propose a self-supervised framework named DASP that leverages spatiotemporal priors for nighttime depth estimation. Specifically, DASP consists of an adversarial branch for extracting spatiotemporal priors and a self-supervised branch for learning. In the adversarial branch, we first design an adversarial network where the discriminator is composed of four devised spatiotemporal priors learning blocks (SPLB) to exploit the daytime priors. In particular, the SPLB contains a spatial-based temporal learning module (STLM) that uses orthogonal differencing to extract motion-related variations along the time axis and an axial spatial learning module (ASLM) that adopts local asymmetric convolutions with global axial attention to capture the multiscale structural information. By combining STLM and ASLM, our model can acquire sufficient spatiotemporal features to restore textureless areas and estimate the blurry regions caused by dynamic objects. In the self-supervised branch, we propose a 3D consistency projection loss to bilaterally project the target frame and source frame into a shared 3D space, and calculate the 3D discrepancy between the two projected frames as a loss to optimize the 3D structural consistency and daytime priors. Extensive experiments on the Oxford RobotCar and nuScenes datasets demonstrate that our approach achieves state-of-the-art performance for nighttime depth estimation. Ablation studies further validate the effectiveness of each component.



\end{abstract}

\section{INTRODUCTION}
\IEEEPARstart{M}{onocular} depth estimation aims to predict dense depth maps from RGB images, which has been widely deployed in various applications, such as 3D scene understanding \cite{weibel2023challenges}, augmented reality \cite{newcombe2011dtam}, and autonomous driving \cite{nuscenes2019}, etc. However, to predict accurate dense depth maps, a large amount of high-quality paired images and depth maps is required, which is tricky to collect from a real-world environment. In this regard, self-supervised methods \cite{zhou2017SfMLearner,yin2018geonet,ref2024Yang} have drawn more attention since they do not require costly ground-truth depth labels and estimate the depth through the inference of geometric cues extracted from monocular videos. Moreover, with the efforts of \cite{zou2018dfnet,godard2019mono2depth}, the performance of self-supervised depth estimation is comparable to supervised methods in multiple scenarios, e.g., KITTI, Cityscapes, etc. Unfortunately, these studies mainly focus on daytime depth estimation, with limited performance when facing challenging nighttime scenes.

\begin{figure}[t]
    \vspace{-2.00mm}
    \setlength{\abovecaptionskip}{-1.25mm}  
	\centering
	\includegraphics[width=0.970\linewidth]{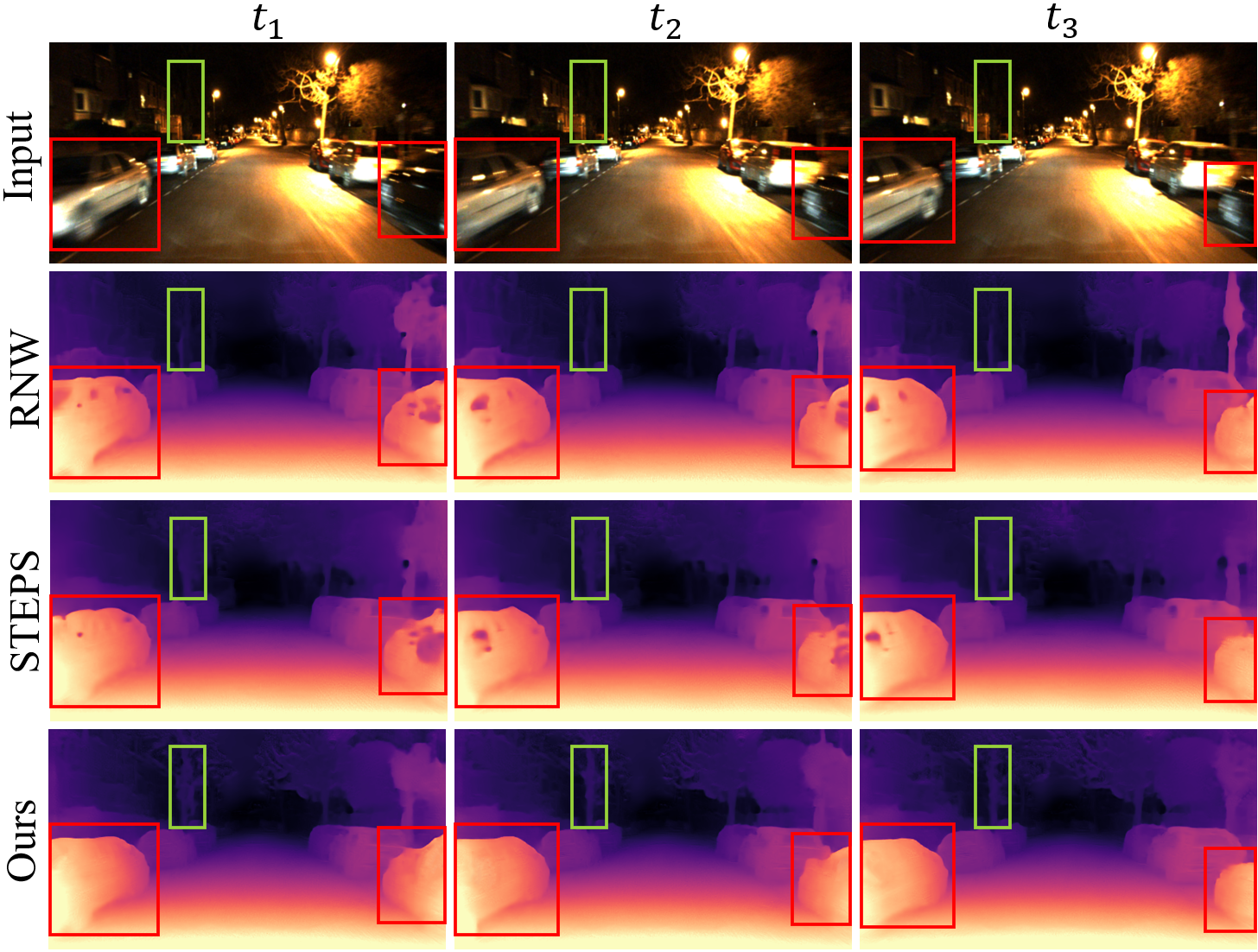}
	\captionsetup{font={small}}
	\caption{The first row shows a set of consecutive image frames from the RobotCar dataset. The next three rows show the depth maps predicted by RNW~\cite{wang2021RNW}, STEPS~\cite{zheng2023steps}, and our method. The green boxes mark a tree, while the red boxes indicate a moving vehicle. From the figures, we can observe that our method effectively captures spatial structure and maintains consistency in dynamic scenes.}
	\vspace{-6.50mm}	
	\label{Fig:1}
\end{figure}

Low visibility and varying illuminations are the main challenges for nighttime depth estimation, which bring a series of problems. For example, low visibility often results in textureless regions that are difficult to recognize, leading to depth missing. As shown in Fig.~\ref{Fig:1}, the green box demonstrates a dark region containing a tree that cannot be accurately captured in the depth map. Although the study in \cite{wang2021RNW} proposed to leverage low-light image enhancement to restore details in low-visibility areas, it still cannot generate accurate depth maps for these areas. Varying illuminations usually happen in the moving objects and streetlights, causing a large area of blur. As shown in Fig.~\ref{Fig:1}, the red box indicates motion blur caused by a moving vehicle, where the car windows are inaccurately estimated. Although several studies \cite{zheng2023steps,cong2024SRNSD} adopted masking mechanisms to bypass dynamic regions, their methods often suffer from inconsistency and instability in these areas. For example, in consecutive frames, the depth of car windows is estimated in the previous frame, but it is lost in the subsequent sequence. 

 
To address these problems, we propose a self-supervised framework named DASP that transfers the spatiotemporal priors from daytime to nighttime for monocular depth estimation. Specifically, the DASP contains an adversarial branch and a self-supervised branch. In the adversarial branch, we first develop an adversarial network where the discriminator is composed of four spatiotemporal priors learning blocks (SPLB). Particularly, the SPLB includes a spatial-based temporal learning module (STLM) to capture motion-related changes along the time axis, and an axial spatial learning module (ASLM) to extract spatial information along orthogonal axes. The integration of STLM and ASLM provides sufficient spatiotemporal features to restore the textureless and blurry regions. In the self-supervised branch, we propose a 3D consistency projection loss that projects the pixels from both the target and source frames to the same 3D coordinate, and computes the discrepancy between frames to enhance spatial consistency and daytime priors. Extensive experiments on the Oxford RobotCar and nuScenes datasets validate the effectiveness and stability of our approach. 

In summary, our main contributions are as follows:

\begin{itemize}
    \item We propose a self-supervised framework that exploits the spatiotemporal representations from daytime priors for guiding nighttime depth estimation.
    
    \item We devise a spatiotemporal priors learning block (SPLB) which consists of two modules: Spatial-based Temporal Learning Module (STLM) and Axial Spatial Learning Module (ASLM). Through the integration of two modules, our model can obtain sufficient spatiotemporal features to restore textureless and blurry regions.
        
    \item We design a 3D projection consistency loss which strengthens the geometric consistency and daytime priors.    
    
    \item Extensive experiments on the Oxford RobotCar and nuScenes datasets demonstrate that our method achieves state-of-the-art performance across multiple metrics.
\end{itemize}

\section{Related Work}

\subsection{Self-supervised Depth Learning from Videos} 
To alleviate the reliance on labeled data, Zhou et al.~\cite{zhou2017SfMLearner} first proposed self-supervised monocular depth estimation by jointly learning depth and pose. This method is designed based on static scenes and cannot deal with dynamic scenes, leading to multi-view ambiguity.
To solve this problem, a number of strategies have been proposed, such as optical flow~\cite{yin2018geonet}, instance segmentation~\cite{casser2019depth-semantic}, uncertainty map~\cite{poggi2020uncertainty}, and stationary pixel mask~\cite{godard2019mono2depth} to recognize moving objects and mask motion regions. Besides, authors in~\cite{li2021unsupervised,lee2021attentive} proposed to model 3D object motion, and authors in~\cite{feng2022disentangling,watson2021manydepth} present to disentangle object motion to construct cost volumes, but these object-level methods lack precise supervision and still remain inherently ambiguous. Recently, Sun et al.~\cite{sun2023Sc-depthv3} leveraged pseudo-depth as depth priors to estimate depth maps and achieved better performance, which verified the effectiveness of geometry priors. 
Based on it, Mono et al.~\cite{moon2024ground} introduced a ground-contacting prior to handle ambiguous moving objects. Although this approach achieves promising results in daytime scenarios, its performance degrades significantly under nighttime conditions.

\subsection{Nighttime Self-supervised Learning Methods}
Considering daytime and nighttime environments have the same structural information, Spencer et al.~\cite{spencer2020defeat} proposed to learn depth-invariant representations from daytime and nighttime. However, their method performed worse in the low-visibility and illumination variability environments. To address these challenges, quite a few methods have been proposed, which can be divided into two categories: domain adaptation and self-distillation. 
In domain adaptation, Vankadari et al.~\cite{ vankadari2020ADFA} and Liu et al.~\cite{liu2021adds} extracted features from daytime and nighttime domains separately, and applied domain adaptation and separation to alleviate the negative effects of poor visibility and uneven lighting. To fully utilize daytime visual cues, Wang et al.~\cite{wang2021RNW} introduced a prior derived from daytime depth distributions to enhance nighttime depth prediction. Zheng et al.~\cite{zheng2023steps} adopted a different approach using image enhancement to adjust exposure and reduce photometric inconsistency between daytime and nighttime. However, their methods still suffer from the smoothness of depth and texture recognition. In this regard, Cong et al.~\cite{cong2024SRNSD} introduced a composite structure regularization strategy that aligns feature and depth output space to ensure multiscale consistency in structural and textural predictions.
In self-distillation, Gasperini et al.~\cite{gasperini2023md4all} adopted GAN to generate adverse samples from daytime images as input, and devised a distillation loss to improve photometric consistency under nighttime conditions. Based on it, Wang et al.~\cite{wang2025promptmono} proposed learnable visual prompts that capture domain-specific knowledge to enhance cross-domain adaptation. Although these methods utilize daytime data as spatial priors to guide nighttime depth estimation, they overlook the photometric consistency along the temporal dimension. In this work, we pretrain a daytime depth model to produce depth sequences as spatiotemporal priors and leverage adversarial learning to enhance temporal and spatial consistency.

\begin{figure*}[htbp]
    \setlength{\abovecaptionskip}{-0.500mm}  
	\centering
	\includegraphics[width=0.900\linewidth]{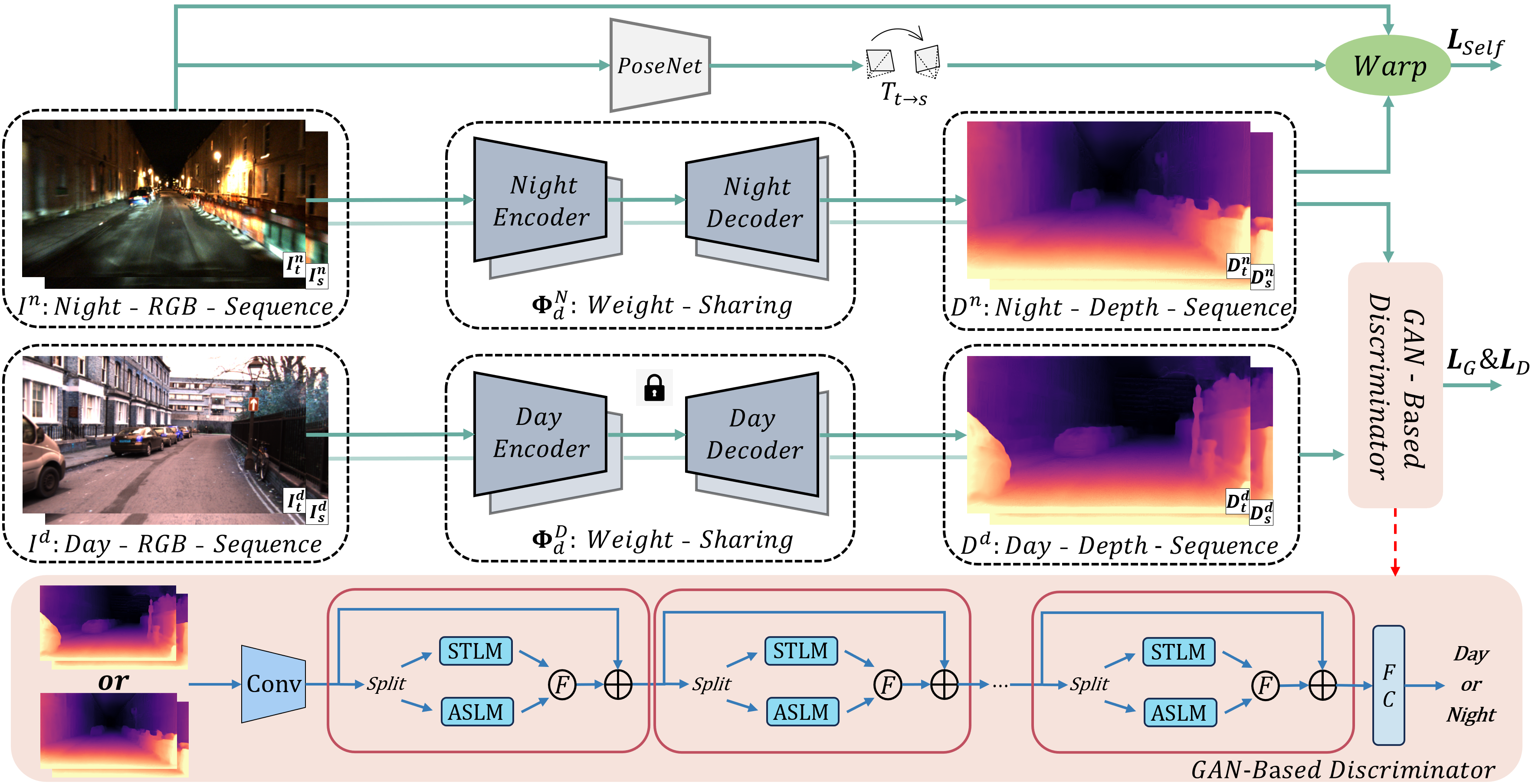}
	\captionsetup{font={small}}
	\caption{Overview of our proposed framework. When given a nighttime sequence $(I^n_t, I^n_s)$, we first use a PoseNet to predict relative pose $T_{t \rightarrow s}$ and an encoder-decoder network to predict depth maps $(D^n_t, D^n_s)$ respectively, and then warp them to construct a geometric mapping for self-supervised learning. While given a daytime sequence $(I^d_t, I^d_s)$, we adopt a pretrained and fixed model to extract depth priors $(D^d_t, D^d_s)$, and together with nighttime depth maps to feed into a GAN-based discriminator with STLM and ASLM to extract spatiotemporal representation and distinguish day and night. Finally, the framework is jointly optimized with self-supervised and adversarial loss.}
	\label{Fig:2}
    \vspace{-5.5mm}
\end{figure*}

\section{Method}
\subsection{Self-supervised Training} \label{sec:3.1}
The task of self-supervised monocular depth estimation is to reproject the pixel $p_t$ in target frame $I_t$ from the pixel ${p}_s$ in source frame $I_s$ through a depth network $\Phi_d$ and a pose network $\Phi_p$, where the depth network predicts the correspondence depth map $D_t=\Phi_d(I_t)$ and the pose network generates the relative pose $T_{t \rightarrow s}= \Phi_p(I_t, I_s)$. The transformation between source pixel $p_s$ and target pixel ${p}_t$ can be formulated as follows:
\begin{equation}
	 p_s \sim K T_{t \rightarrow s} D_t(p_t) K^{-1} {p}_t
        \label{equ: Geometric}
\end{equation}
where $\sim$ denotes the homogeneous equivalence and $K$ represents the camera intrinsic matrix. Based on the transformation, the target frame $\hat I_t$ can be recovered from $I_s$ by:
\begin{equation}
	\hat I_t=\langle I_s, p_s \rangle
\end{equation}
where $ \langle \cdot \rangle $ denotes the differentiable bilinear sampling. 
Intuitively, to reduce the reprojection error, we first follow \cite{wang2004image} to introduce a photometric consistency loss that combines a weighted SSIM and weighted $\ell_1$ error to calculate the difference between $I_t$ and $\hat I_t$:
\begin{equation}
    \mathcal{L}_{\text{p}} = \alpha \cdot \frac{1 - \text{SSIM}(I_t, \hat{I}_t)}{2} + (1 - \alpha) \cdot \| I_t - \hat{I}_t \|_1
\end{equation}
where the weight $\alpha$ is set to 0.85. After that, we follow \cite{godard2017monodepth} to apply a disparity smoothness loss to facilitate the smoothness of generated depth and avoid depth ambiguity:
\begin{equation}
	\mathcal{L}_{\text{ds}} = \lvert \partial_x D_t \rvert e^{-\lvert \partial_x I_t \rvert} + \lvert \partial_y D_t \rvert e^{-\lvert \partial_y I_t \rvert}
\end{equation}

\begin{figure}[tb]
    \setlength{\abovecaptionskip}{-0.75mm}  
	\centering
    \includegraphics[width=0.70\linewidth]{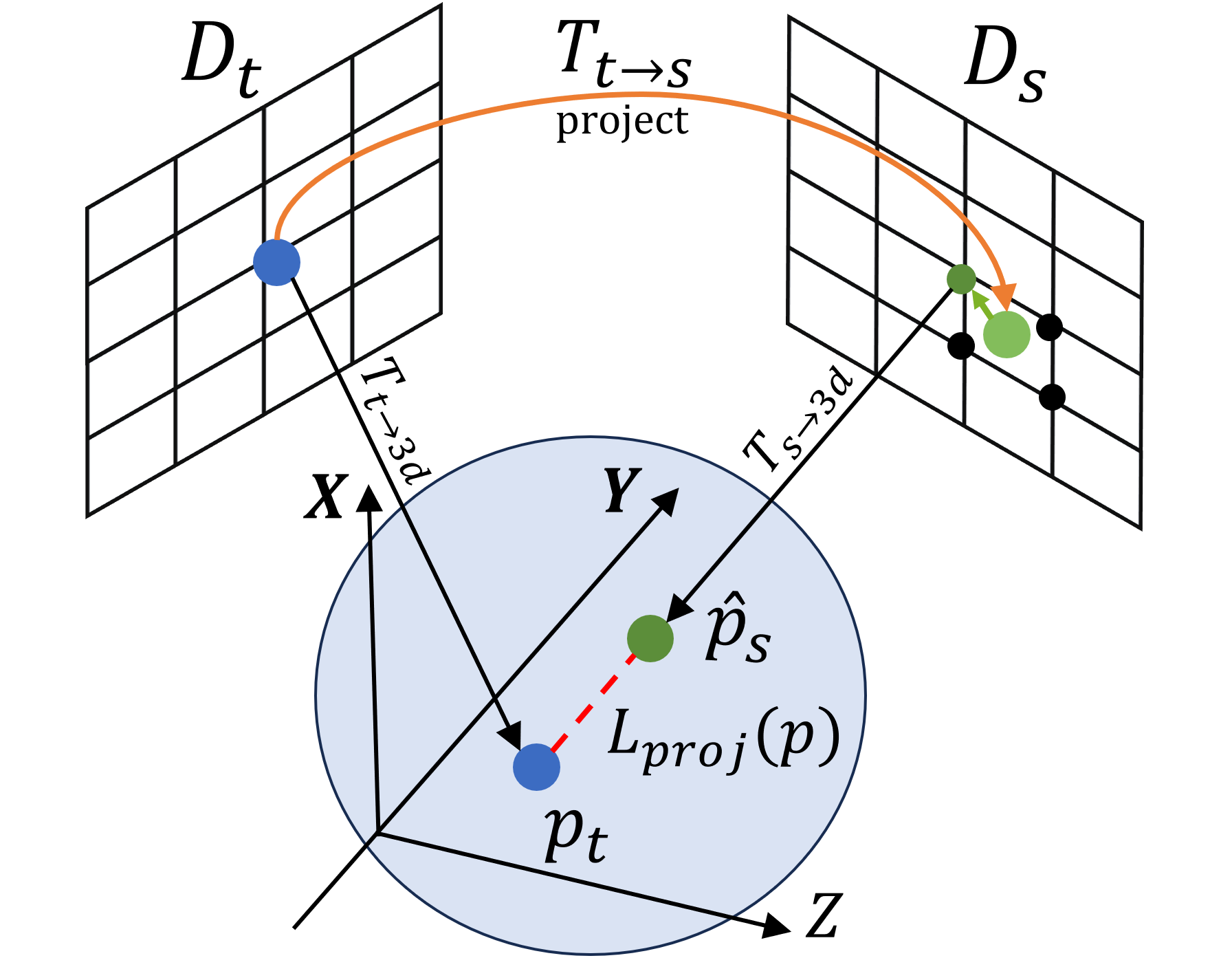}
	\captionsetup{font={small}}
	\caption{The computation of 3D projection consistency loss. According to Eq.~\ref{equ: Geometric}, we first reproject the target point $p_t$ in the target depth map $D_t$ to the source depth map $D_s$ to obtain the interpolated point $\hat{p}_s$. And then two pose networks are deployed to project the points into a shared 3D coordinate. Finally, we calculate the Euclidean distance between two points as a 3D projection consistency loss.}
	\vspace{-6.15mm}
	\label{Fig:3}
\end{figure}

Furthermore, we adopt the geometric consistency loss \cite{bian2019unsupervised} to penalize depth inconsistencies between adjacent frames:
\begin{equation}
	\mathcal{L}_{\text{geom}} = \frac{1}{\lvert V \rvert} \sum_{p \in V} D_{\text{diff}}(p)
\end{equation}
where $V$ is the set of valid projections within image boundaries, and $D_{\text{diff}}$ measures per-pixel depth inconsistency between $D_t$ and $D_s$. This yields a self-discovered mask:
\begin{equation}
	M_{s} = 1-D_\text{diff}
\end{equation}
where $M_s \in [0, 1]$ highlights the view-consistent regions while suppressing inconsistent parts. 

Although previous works take pixel-wise error and smoothness into consideration, they only focus on frame-level error, i.e., they only compute the unidirectional reprojection from the target frame to the source frame without considering the spatial diversity during reprojection, which makes the model extremely unstable when dealing with occluded or blurred areas. To this end, we design a 3D projection consistency loss that projects the pixel from both the source and target frames into the shared 3D space, and then computes the discrepancy between the two projected pixels, thus optimizing the depth network and pose network. Fig.~\ref{Fig:3} illustrates the projection process, and the loss is formulated as follows:
\begin{equation}
    \mathcal{L}_{\text{proj}} = \left\| D_s(\hat{p}_s) K^{-1} \hat{p}_s - T_{t \rightarrow s} D_t(p_t) K^{-1} {p}_t \right\|_2
    \label{equ:Projection}
\end{equation}
where $D_s$ and $D_t$ represent the depth map predicted by depth network, $\hat p_s$ is the reprojected pixel after differentiable bilinear sampling.
Through the 3D projection consistency loss, the discrepancy between two projected points can be calculated via depth, which is more direct and intuitive.

Finally, the total self-supervised loss is a weighted combination of the aforementioned losses:
\begin{equation}
	\mathcal{L}_{\text{self}} = \lambda_1 \mathcal{L}_{\text{p}} \otimes M_s + \lambda_2 \mathcal{L}_{\text{ds}} + \lambda_3 \mathcal{L}_{\text{geom}} + \lambda_4 \mathcal{L}_{\text{proj}}
    \label{equ:6}
\end{equation}
where $\lambda_{\text{1}},\lambda_{\text{2}},\lambda_{\text{3}}$ and $\lambda_{\text{4}}$ are empirically set as 0.7, 0.1, 0.5, and 0.5, respectively.

\begin{figure*}[htbp]
    \setlength{\abovecaptionskip}{-0.53mm}  
	\centering
    \includegraphics[width=0.88\linewidth, height=6.65cm]{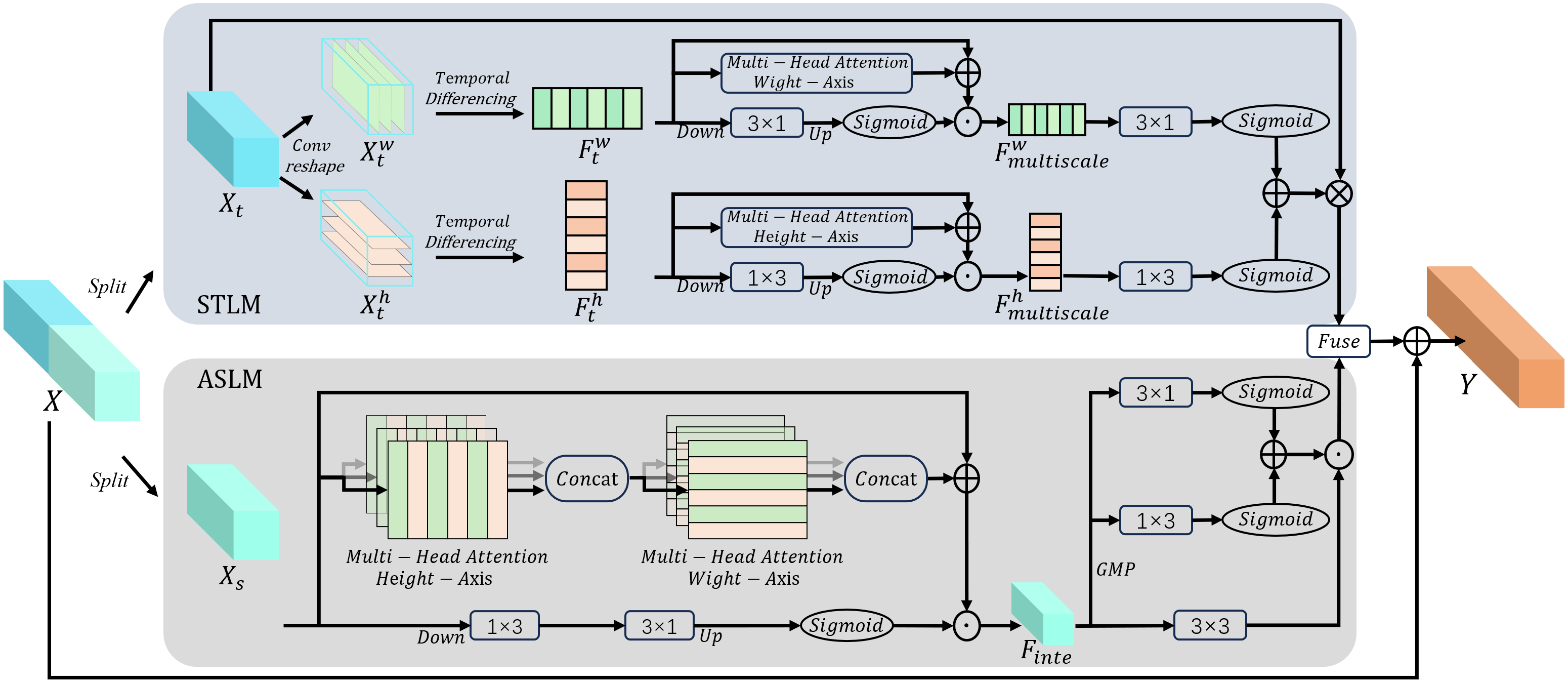}
	\captionsetup{font={small}}
        \caption{Overview of spatiotemporal priors learning block (SPLB). First of all, the input ${X}$ is split into two parts, ${X}_s$ and ${X}_t$, and then they are sent to the Spatial-based temporal learning module (STLM) and the Axial spatial learning module (ASLM), respectively. The STLM captures temporal features via orthogonal differencing of adjacent frames, while the ASLM uses asymmetric convolutions with global axial attention to extract spatial features. Finally, the outputs of the two modules are fused to generate the final spatiotemporal representation ${Y}$.}
        \vspace{-5.8mm}	
        \label{Fig:4}
\end{figure*}

\vspace{-1.5mm}	
\subsection{Adversarial Adaptation to Learn Spatiotemporal Priors} \label{sec:3.2}
Considering that GANs have been proven to learn the underlying patterns from data, we adopt a GAN-based network to extract spatiotemporal features from daytime priors. Specifically, for the generator, we use Monodepth2~\cite{godard2019mono2depth} to generate indistinguishable nighttime samples. Additionally, we pretrain another Monodepth2 on daytime data to provide daytime priors.
For the discriminator, we stack four elaborated Spatiotemporal Priors Learning Blocks (SPLB), each of which consists of two branches, STLM and ASLM, to extract appropriate temporal and spatial depth representation patterns from input $X$. Fig.~\ref{Fig:4} shows the overview of the SPLB.

\textbf{1) Spatial-based temporal learning module (STLM).} In video sequences, temporal priors could directly guide spatial structure. For example, in the dynamic scenes, previous frames could provide more accurate spatial priors for the subsequent frames. To exploit this spatiotemporal consistency, we devise STLM, which decomposes the entire sequence along the time axis and applies convolutional differences to capture motion-related changes between consecutive frames.

Specifically, given a temporal input \( X_t \in \mathbb{R}^{T \times \frac{C}{2} \times H \times W} \), we first compressed it by a factor of $r$ in the channel dimension and reshaped it into a temporal sequence along the horizontal and vertical axes. Because the horizontal branch and vertical branch adopt the same process, we take the horizontal axis as an example here:
\begin{equation}
X_t^w \in \mathbb{R}^{H \times \frac{C}{2r} \times W \times T}, \quad X_t^w = \{x_1^w, x_2^w, \dots, x_T^w\}
\end{equation}
where each \( x_t^w \in \mathbb{R}^{H \times \frac{C}{2r} \times W} \) denotes the horizontal features at time $t$. And then we deploy a convolution network with zero-padding to compute directional differences along the time dimension to obtain inter-frame motion features:
\begin{equation}
F_t^w = \zeta_{3 \times 3} (x_{t+1}^w) - x_t^w, \quad t = 1, \dots, T-1
\end{equation}
where $\zeta$ denotes the convolutions, and the subscript indicates the kernel sizes. After that, a three-branch structure with axis-specific asymmetric convolution, axial attention, and a residual connection is used to capture multiscale inter-frame variations:
\begin{equation}
        F^w_{local} = \sigma(Up (\zeta_{3 \times 1} (Down (F^w_t))))
\end{equation}
\begin{equation}
		F^w_{multiscale} = F^w_{local} \odot (F^w_t + MHA_W(F^w_t) )
\end{equation}
where $\sigma$ is the sigmoid function, $F_{local}^w$ and ${MHA}_W$ denote the local features and global attention \cite{wang2020axial} along the width axes. To further refine the direction-aware temporal, we apply an activated asymmetric convolution with a sigmoid function:
\vspace{-2mm}
\begin{equation}
    F^w_{\text{refine}} = \sigma(\zeta_{1\times3}(F^w_{\text{multiscale}}))
\end{equation}

Finally, we combine two refined axes features and the original time input $X_t$ to generate the final temporal features:

\begin{equation}
  \raisebox{0.1ex}{$F_{t} = (F^w_{refine} + F^h_{refine}) \odot X_t$}
\end{equation}
where $F^w_{refine}$ and $F^h_{refine}$ represent the refined horizontal and vertical features, respectively.

\captionsetup[table]{
  labelformat=simple,
  labelsep=newline,
  name=TABLE, 
  textfont=sc,
  justification=centering,
  font=scriptsize
}

\begin{table*}[htbp]
	\centering
	\renewcommand{\arraystretch}{0.97}
	\setlength{\tabcolsep}{10.5pt}
	\caption{Comparison with state-of-the-art methods, where lower error and higher accuracy indicate better performance.} 
    \vspace{-1.0mm}
	\begin{tabular}{c|c|cccc|ccc}
		\hline
		\multirow{2}{*}{\textbf{Methods}} & \multirow{2}{*}{\textbf{Max Depth}} 
		& \multicolumn{4}{c|}{\textbf{Error} $\downarrow$} & \multicolumn{3}{c}{\textbf{Accuracy} $\uparrow$} \\
		\cline{3-9}
		& & \textbf{Abs Rel}	& \textbf{Sq Rel} & \textbf{RMSE} & \textbf{RMSE log} & \raisebox{-0.3ex}{$\delta < 1.25$} & \raisebox{-0.3ex}{$\delta < 1.25^2$} & \raisebox{-0.3ex}{$\delta < 1.25^3$} \\ 
		\hline
		\multicolumn{9}{c}{\textbf{RobotCar-Night}} \\
		\hline
		Monodepth2\cite{godard2019mono2depth} & 40m & 0.661 & 25.213 & 12.187 & 0.553 & 0.551 & 0.849 & 0.914 \\
        ADDS\cite{liu2021adds} & 40m  & 0.233 & 2.344 & 6.859 & 0.270 & 0.631 & 0.908 & 0.962 \\
        md4all-DD\cite{gasperini2023md4all} & 40m & 0.202  & 1.882 & 7.929 & 0.264 & 0.642 & 0.921 &0.970 \\
        ACDepth\cite{ACDepth} & 40m  & 0.187 & 1.633 & 6.843 & 0.242 & 0.703 & 0.925 & 0.971 \\
        PromptMono\cite{wang2025promptmono} & 40m & 0.206 & 2.057 & 6.497 & 0.246 & 0.736 & 0.917 & 0.966 \\
		RNW\cite{wang2021RNW}  & 40m  & 0.176 & 1.323 & 4.922 & 0.225 & 0.772 & 0.933 & 0.975 \\
		STEPS\cite{zheng2023steps} & 40m  & 0.154 & 1.108 & 4.682 & 0.213 & 0.803 & 0.937 & 0.974 \\
		SRNSD\cite{cong2024SRNSD} & 40m  & 0.136 & 0.799 & 4.257 & 0.194 & 0.836 & 0.951 & 0.983 \\
		\textbf{Ours} & 40m & \textbf{0.132} & \textbf{0.786} & \textbf{4.125} & \textbf{0.187} & \textbf{0.849} & \textbf{0.958} & \textbf{0.988} \\
		\hline
		Monodepth2\cite{godard2019mono2depth} & 60m & 0.580 & 21.446 & 12.771 & 0.521 & 0.552 & 0.840 & 0.920 \\
		ADDS\cite{liu2021adds} & 60m  & 0.231 & 2.674 & 8.800 & 0.286 & 0.620 & 0.892 & 0.956 \\
        md4all-DD\cite{gasperini2023md4all} & 60m  & 0.206 & 2.066 & 7.790 & 0.262 & 0.669 & 0.910 & 0.967 \\
        ACDepth\cite{ACDepth} & 60m  & 0.198 & 1.921 & 7.372 & 0.255 & 0.681 & 0.913 & 0.963 \\
		RNW\cite{wang2021RNW}  & 60m  & 0.185 & 1.894 & 7.319 & 0.246 & 0.735 & 0.910 & 0.965 \\	
		STEPS\cite{zheng2023steps} & 60m  & 0.170 & 1.686 & 6.797 & 0.234 & 0.758 & 0.923 & 0.968 \\
        PromptMono\cite{wang2025promptmono} & 60m  & 0.172 & 1.540 & 6.567 & 0.233 & 0.763 & 0.924 & 0.972 \\
		SRNSD\cite{cong2024SRNSD} & 60m  & 0.169 & 1.450 & 6.439 & 0.226 & 0.768 & 0.926 & 0.975 \\
		\textbf{Ours} & 60m & \textbf{0.164} & \textbf{1.442} & \textbf{6.315} & \textbf{0.218} & \textbf{0.777} & \textbf{0.930} & \textbf{0.981} \\
		\hline
		\multicolumn{9}{c}{\textbf{NuScense-Night}} \\
		\hline
		Monodepth2\cite{godard2019mono2depth} & 60m  & 1.185 & 42.306 & 21.613 & 1.567 & 0.184 & 0.360 & 0.504 \\
		RNW\cite{wang2021RNW} & 60m  & 0.326 & 3.999 & 9.932 & 0.417 & 0.492 & 0.765 & 0.870 \\
        Light-Dark\cite{liang2024lightdark} & 60m  & 0.340 & 4.838 & 10.136 & 0.414 & 0.526 & 0.772 & 0.889 \\
		STEPS\cite{zheng2023steps} & 60m  & 0.292 & 3.363 & 9.120 & 0.390 & 0.572 & 0.805 & 0.908 \\
		\textbf{Ours} & 60m & \textbf{0.276} & \textbf{3.072} & \textbf{8.819} & \textbf{0.367} & \textbf{0.584} & \textbf{0.809} & \textbf{0.916} \\
		\hline
	\end{tabular}
	\label{table:1}
    \vspace{-2.5mm}	
\end{table*}

\textbf{ 2) Axial spatial learning module (ASLM).} Since the street scene images are captured by the cameras and LiDARs on the car, the view of street scenes extends vertically from near to far, while the depth decreases horizontally from near to far. Besides, street scenes often contain many structural objects, such as streetlights, buildings, and cars, which generally follow the distribution of vertical and horizontal axes. Based on these observations, we proposed ASLM, which uses local asymmetric convolutions with global axial attention to extract multiscale structural depth representations and leverage them to guide depth estimation at nighttime.

Specifically, given an input $X_s \in \mathbb{R}^{T \times \frac{C}{2} \times H \times W}$, we apply multi-head self-attention along the height and width axes to extract global structural features:
\vspace{-1mm}
\begin{equation}
	F_{global} = MHA_W(MHA_H(X_s))
\end{equation} 
where ${MHA}_H$ denotes attention along the height axes. Additionally, to preserve original spatial information, a residual connection is adopted. Considering attention mechanism pays more attention to the global contextual features, we apply two asymmetric convolutions in input $X_s$ to obtain local features and multiply them with the attention features to obtain the integrated features $F_{inte}$:

\begin{equation}
    F_{local} =  \sigma(\operatorname{Up}(\zeta_{3\times 1} (\zeta_{1 \times 3} (\operatorname{Down}(X_s))))) \\
\end{equation}
\begin{equation}
     F_{inte} = F_{local} \odot (F_{global} + X_s)
\end{equation}

To further refine direction-aware features, we apply global max pooling on $F_{inte}$ and deploy two asymmetric convolutions to compute horizontal and vertical attention maps and add them together:
\vspace{-0.45mm}
\begin{equation}
    F_{dire} = \sigma(\zeta_{3\times1}(GMP(F_{inte}))) + \sigma(\zeta_{1\times3}(GMP(F_{inte})))
\end{equation}
where $F_{dire}$ is direction-aware features. Finally, we use a $3\times3$ convolution to extract refined features,  and multiply with the $F_{dire}$ to obtain the final spatial features $F_s$:
\vspace{-0.45mm}
\begin{equation}
F_s = \zeta_{3 \times 3} ( F_{inte}) \odot F_{dire}
\end{equation}

\textbf{3) Integration of spatiotemporal features.} 
To integrate temporal and spatial features, we adopt a $ 3 \times 3 $ convolution to fuse the $F_t$ and $F_s$:
\vspace{-0.75mm}
\begin{equation}
Y_{inte} = \zeta_{3 \times 3} (F_t + F_s)
\end{equation}
where $Y_{inte}$ is the integrated features. We add it to two branches, and then concatenate them along the channel dimension. The final output $Y$ is formed by concatenating the features and the original input $X$ via a residual connection:
\vspace{-0.55mm}
\begin{equation}
Y = \text{Concat}(F_t + Y_{inte},\, F_s + Y_{inte}) + X
\end{equation}

\begin{figure*}[htbp]
    \setlength{\abovecaptionskip}{0.15mm}
	\centering  
	\includegraphics[width=0.915\linewidth]{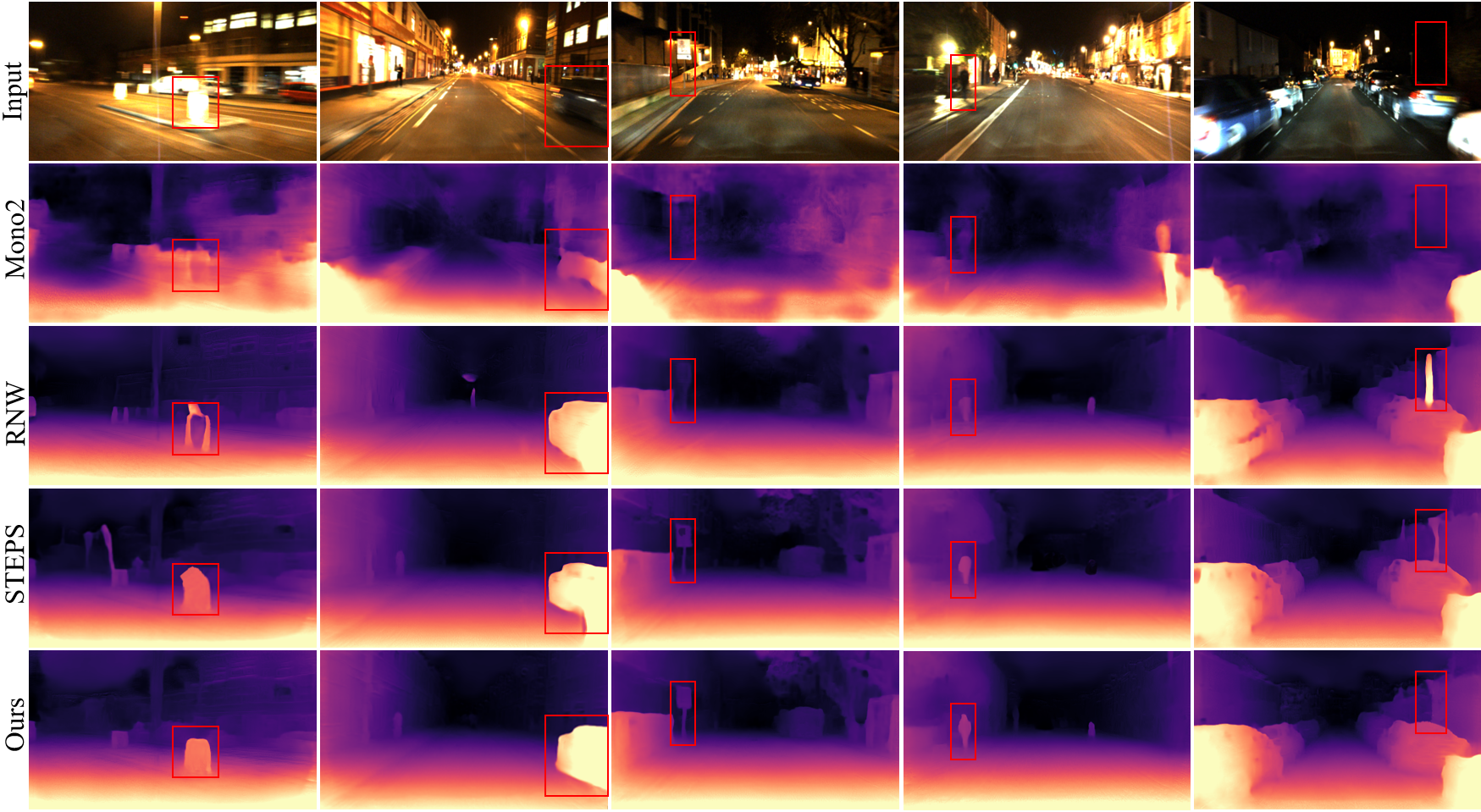}
	\captionsetup{font={small}}
        \caption{Qualitative comparison of state-of-the-art methods on the RobotCar dataset, where the key differences are highlighted with red boxes.}
	\label{Fig:5}
    \vspace{-4.5mm}	
\end{figure*}

\vspace{-2.5mm}	
\subsection{Final loss} \label{sec:3.3}
In order to optimize the depth maps $D^n$ generated by the nighttime generator $\Phi^N_d$, we introduce a pretrained daytime model $\Phi^D_d$ to generate accurate depth maps $D^d$ as priors to confuse the discriminator $\Phi_A$ and force the nighttime generator to mimic. The loss function is as follows:
\begin{gather}
   	\mathcal{L}_\text{D} = \frac{1}{2 N^d } \sum_{D^d} (\Phi_A(D^d)-1)^2 
   	+ \frac{1}{2 N^n } \sum_{D^n} (\Phi_A(D^n))^2 \label{equ:19} \\[-2pt]
   	\mathcal{L}_\text{G} = \frac{1}{2 N^n } \sum_{D^n} (\Phi_A(D^n)-1)^2 \label{equ:20}
\end{gather}
where $N^d$ and $N^n$ are the number of daytime and nighttime training images. Note that the depth maps here are not fixed to two frames, but refer to a sequence of depth maps.

In summary, the final loss is composed of Self-supervised loss, generator loss, and discriminator loss: 
\vspace{-0.5mm}	
\begin{equation}
\mathcal{L}_\text{total} = \mathcal{L}_{\text{self}}  + \mathcal{L}_{\text{G}} + \mathcal{L}_{\text{D}} \label{equ:21}
\end{equation}

\section{EXPERIMENT}
\subsection{Dataset}
\textbf{RobotCar.} Oxford RobotCar \cite{robotcar2017} is a large-scale urban driving dataset under diverse conditions. We build RobotCar-Night using left images from the front stereo camera in sequence 2014-12-16-18-44-24, cropped to $1152 \times 672$. The training set has 19k frames from the first five splits (excluding stationary frames), and the test set has 411 frames from the fifth and sixth splits. Depth GT for testing is generated using the official toolbox with front LMS LiDAR and INS data.

\textbf{nuScenes.} nuScenes \cite{nuscenes2019} contains 1000 driving scenes in Boston and Singapore. We select 60 nighttime scenes, crop images to $1536 \times 768$, and use over 10k frames for training and 500 for testing. Test depth GT is obtained from top LiDAR via the official toolbox.


\subsection{Implementation Details}
The daytime depth estimation network is pretrained based on Monodepth2~\cite{godard2019mono2depth} to generate spatiotemporal priors, and the nighttime network is optimized by self-supervised training and adversarial learning. The network was trained for 50 epochs on an RTX 3090 GPU using the Adam optimizer with a batch size of 8. We set the initial learning rate as $3e^{-5}$, linearly warmed up to $1\times10^{-4}$ after 500 iterations, and halved at the 15th epoch. To verify the efficiency of our method, we also perform the experiments on an embedded platform NVIDIA Jetson AGX Orin. In terms of model evaluation, we choose seven standard metrics, including: Abs Rel, Sq Rel, RMSE, RMSE log, and accuracy with the thresholds $\delta$ of 1.25, $1.25^2$, and $1.25^3$. Notably, although multi-frame information is utilized during training, only a single frame is required at inference time. For comparison, we compare our method with state-of-the-art monocular nighttime depth estimation approaches, including ADDS \cite{liu2021adds}, RNW \cite{wang2021RNW}, STEPS \cite{zheng2023steps}, md4all-DD \cite{gasperini2023md4all}, Light-Dark \cite{liang2024lightdark}, SRNSD \cite{cong2024SRNSD}, PromptMono \cite{wang2025promptmono}, and ACDepth \cite{ACDepth}. All the results are reported under the depth ranges of 40m and 60m, and all comparison approaches are trained and tested on the same dataset.

\begin{figure*}[htbp]
    \setlength{\abovecaptionskip}{-0.95mm}
	\centering
	\includegraphics[width=0.9230\linewidth]{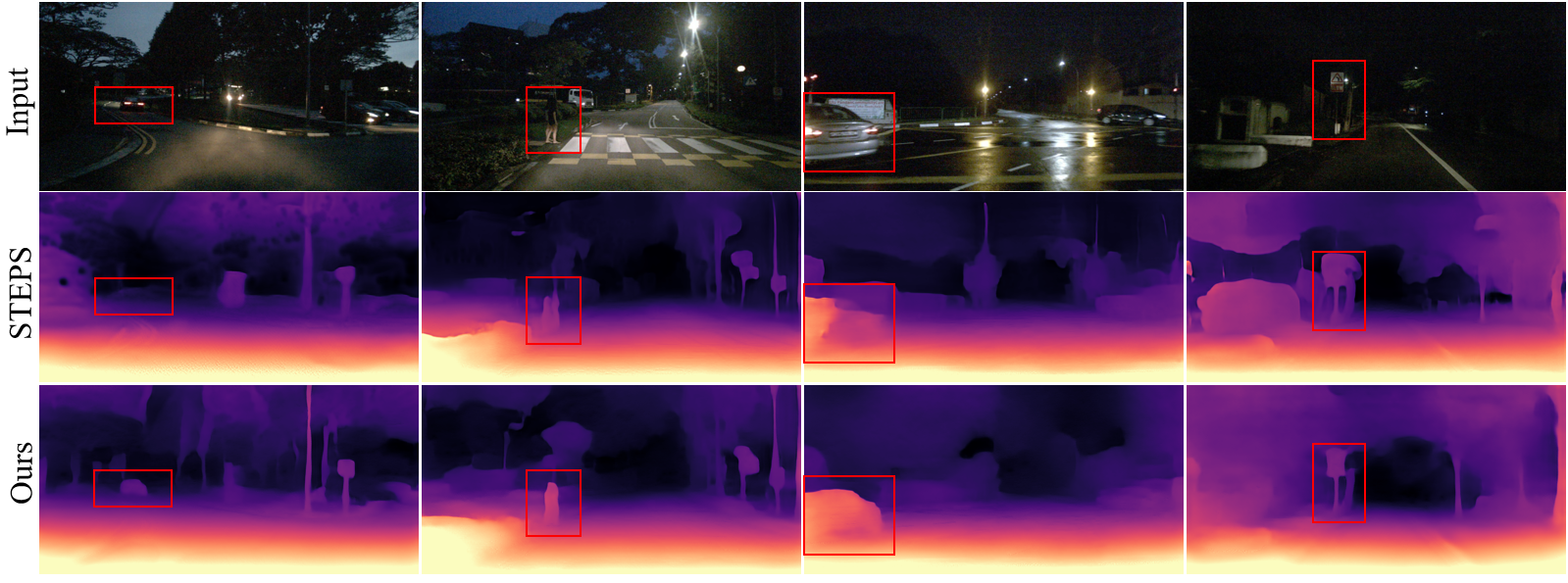}
	\captionsetup{font={small}}
        \caption{Qualitative comparison of state-of-the-art methods on the nuScenes dataset, where the key differences are highlighted with red boxes.}
	\label{Fig:6}
        \vspace{-4.0mm}	
\end{figure*}

\vspace{-2.0mm}	
\subsection{Compare with State-of-the-art Methods}
Table \ref{table:1} summarizes a quantitative comparison of the state-of-the-art approaches on the Oxford RobotCar dataset \cite{robotcar2017} and the nuScenes dataset \cite{nuscenes2019}. From the table, we can observe that daytime-oriented methods such as Monodepth2 \cite{godard2019mono2depth}, which is retrained on nighttime data, perform poorly at night. While recent domain adaptation-based methods and self-distillation approaches achieve notable performance improvements. Compared with state-of-the-art methods, our method outperforms other methods on all metrics in the range of 40m and 60m, demonstrating its effectiveness and robustness. In particular, on the RobotCar dataset, our network achieves 2.94\% and 2.96\% improvements in Abs Rel over the SRNSD in the range of 40m and 60m; while on the nuScenes dataset, our method achieves a 5.48\% improvement in Abs Rel compared with STEPS in the range of 60m. We believe it is beneficial to the proposed SPLB that captures the spatiotemporal priors to guide depth estimation and the 3D projection consistency loss to maintain consistency in 3D space. In terms of efficiency, the inference times based on RTX 3090 and NVIDIA Jetson AGX Orin are 163.9 fps and 36.6 fps, respectively.

Intuitively, we visualize several depth estimation examples from RobotCar and nuScenes datasets, where the key differences are highlighted with red boxes. Fig. \ref{Fig:5} reveals the examples from RobotCar, from which we can observe that all the methods can well estimate the road surface, but when handling the moving objects and photometrically inconsistent areas, other methods estimate the wrong depth maps or produce inconsistent depth values. In contrast, our proposed method is able to generate clear and smooth depth maps thanks to the SPLB that extracts the spatial features of structural objects (such as guideposts, tree trunks, etc.) and temporal priors of the moving objects (such as cars, pedestrians, etc.). Fig. \ref{Fig:6} demonstrates the examples from the more challenging nuScenes dataset, from which we can notice that under low illuminated and noise-corrupted environments, our model is still able to produce accurate depth maps with clear contours and well-preserved structures, e.g., the clear moving car and the distinct guidepost. We believe it is because the 3D projection consistency loss bridges the objects from the target and source frames in a shared 3D space, facilitating the estimation of the objects.

\begin{figure}[t]
	\vspace{1.0mm}
    \setlength{\abovecaptionskip}{-0.70mm}
	\centering
	\includegraphics[width=0.915\linewidth]{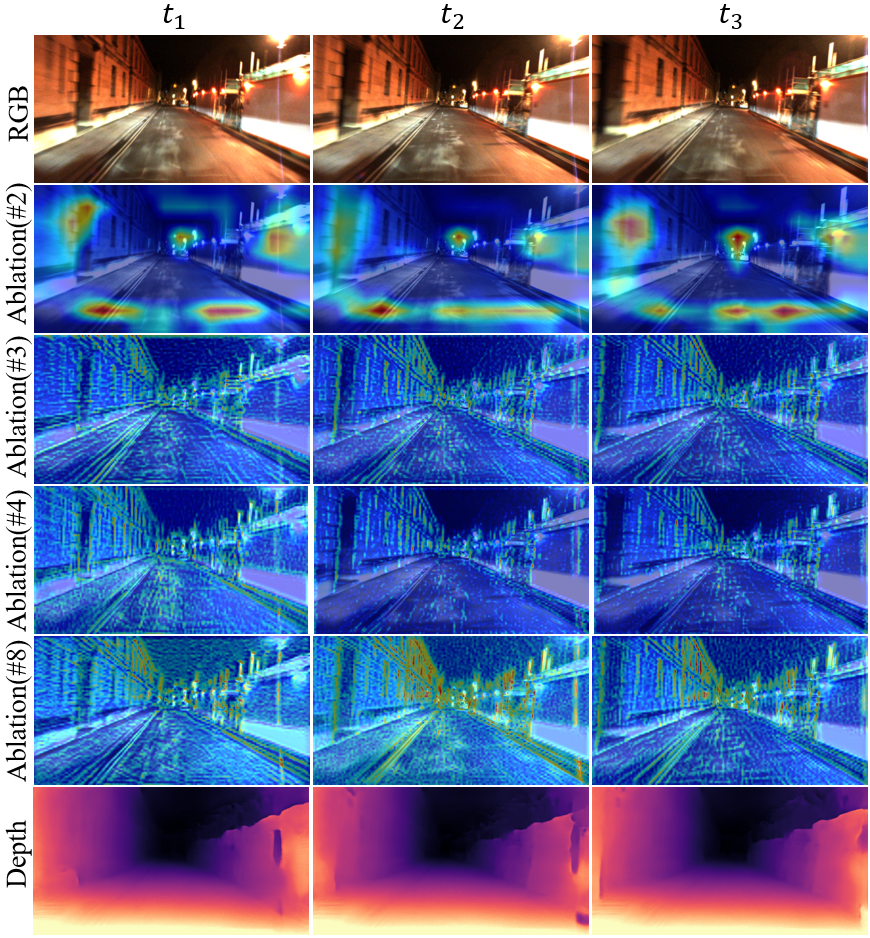}
	\captionsetup{font={small}}
	\caption{Visualization of the ablation study.}
	\vspace{-6.9mm}		
	\label{Fig:7}
\end{figure}

\begin{table*}[htbp]
    \setlength{\abovecaptionskip}{0.30mm}
	\centering
	\renewcommand{\arraystretch}{1.01}
	\setlength{\tabcolsep}{10.5pt}
	\caption{Quantitative results on Robotcar dataset. The depth range is set to 60m, and the best results are marked in bold.}
	\begin{tabular}{c|c|c|c|cccc|ccc}
		\hline
		\multirow{2}{*}{\textbf{\#}} & \multirow{2}{*}{\textbf{Proj   }} & \multicolumn{2}{c|}{\textbf{SPLB}} 
		& \multicolumn{4}{c|}{\textbf{Error} $\downarrow$} & \multicolumn{3}{c}{\textbf{Accuracy} $\uparrow$} \\
		\cline{3-11}
		& & \textbf{STLM} & \textbf{ASLM} & \textbf{Abs Rel}	& \textbf{Sq Rel} & \textbf{RMSE} & \textbf{RMSE log} & \raisebox{-0.25ex}{$\delta < 1.25$} & \raisebox{-0.25ex}{$\delta < 1.25^2$} & \raisebox{-0.25ex}{$\delta < 1.25^3$}  \\
		\hline
        1 & $\times$ & $\times$ & $\times$ & 0.178  & 1.758 & 7.251 & 0.244 & 0.742 & 0.917 & 0.970 \\
        2 & $\checkmark$ & $\times$ & $\times$ & 0.176 & 1.721 & 7.147 & 0.239 & 0.747 & 0.920 & 0.971 \\
        3 & $\times$ & $\checkmark$ & $\times$ & 0.172 & 1.646 & 7.083 & 0.235 & 0.758 & 0.922 & 0.974 \\
        4 & $\times$ & $\times$ & $\checkmark$ & 0.173 & 1.653 & 7.009 & 0.234 & 0.757 & 0.921 & 0.974 \\
        5 & $\times$ & $\checkmark$ & $\checkmark$ & 0.168 & 1.499 & 6.830 & 0.228 & 0.767 & 0.924 & 0.977 \\
        6 & $\checkmark$ & $\checkmark$ & $\times$ & 0.167 & 1.486 & 6.722 & 0.225 & 0.769 & 0.926 & 0.977 \\
		7 & $\checkmark$ & $\times$ & $\checkmark$ & 0.168 & 1.493 & 6.859 & 0.226 & 0.767 & 0.924 & 0.976  \\
		8 & $\checkmark$ & $\checkmark$ & $\checkmark$ & \textbf{0.164} & \textbf{1.442} & \textbf{6.315} & \textbf{0.218} & \textbf{0.777} & \textbf{0.930} & \textbf{0.981}\\
		\hline
	\end{tabular}
    \vspace{-6.5mm}	
	\label{table:3}
\end{table*} 

\vspace{-2.0mm}	
\subsection{Ablation Study}


\textbf{Importance of the 3D Projection Consistency Loss.} 
By comparing $\#1$ and $\#2$ in Table~\ref{table:3}, we can find that the 3D projection consistency loss can improve certain performance, indicating its effectiveness in enhancing geometric consistency through the alignment of depth predictions in a shared 3D space and optimizing the daytime prior.

\textbf{Effectiveness of Spatiotemporal Priors Learning Block (SPLB).} 
The SPLB consists of two submodules, the spatial-based temporal learning module (STLM) and the axial spatial learning module (ASLM), so we conduct three ablation studies to verify the effectiveness of these submodules. Specifically, by comparing $\#2$ and $\#6$, as well as $\#2$ and $\#7$ in Table~\ref{table:3}, we find that both STLM and ASLM have a critical impact on the nighttime depth estimation since STLM captures motion-related variations along the time axis and ASLM extracts the spatial patterns along the orthogonal axis. When combining STLM and ASLM (as shown in $\#5$), our model achieves the best results, effectively demonstrating the advantage of the dual-branch design and the complementary effect of temporal and spatial features in depth estimation.

\textbf{Effectiveness of the combination of 3D Projection Consistency Loss and Spatiotemporal Priors Learning Block (SPLB).}
By comparing $\#2$, $\#3$, $\#4$, $\#5$, and $\#8$, we can notice that the improvements are limited when only deploying 3D Projection Consistency Loss or SPLB. However, by combining 3D Projection Consistency Loss and SPLB, the results are significantly improved. We believe this is because the 3D Projection Consistency Loss, as a self-supervised loss, requires sufficient spatial-temporal features and structural information for self-learning, and STLM and ASLM precisely provide the spatial-temporal priors to the 3D Projection Consistency Loss, thereby strengthening the performance of the network.

\textbf{Visualization and Analysis.}   
To further investigate the effectiveness of each proposed component, we utilize Grad-CAM to visualize the attention maps for each component, including the 3D Projection Consistency Loss ($\#2$), STLM ($\#3$), ASLM ($\#4$), and all components ($\#8$) with three consecutive frames. As shown in Fig.~\ref{Fig:7}, $\#2$ can maintain attention across frames, but lacks structural information.
$\#3$ captures the motion-related changes and maintains consistency along the time axis. $\#4$ provides clear spatial information, but cannot maintain temporal consistency. While $\#8$ can produce more stable and concentrated attention along axes and maintain consistent attention on 3D structure over time. These indicate that the 3D projection consistency loss can maintain prior consistency, and the STLM and ASLM can effectively capture accurate spatiotemporal representation.

\begin{table}[tb]
	\centering
    \setlength{\abovecaptionskip}{0.40mm}
	\renewcommand{\arraystretch}{1.10}
	\setlength{\tabcolsep}{2.5pt}
	\caption{Comparison of the number of spatiotemporal prior frames.}
	\begin{tabular}{c|ccc|ccc}
		\hline
		 \textbf{Number of} 
		& \multicolumn{3}{c|}{\textbf{Error} $\downarrow$} & \multicolumn{3}{c}{\textbf{Accuracy} $\uparrow$} \\
		\cline{2-7}
		\textbf{Frames} & \raisebox{-0.25ex}{\textbf{Abs Rel}}& \raisebox{-0.25ex}{\textbf{Sq Rel}} & \raisebox{-0.25ex}{\textbf{RMSE}} & \raisebox{-0.27ex}{$\delta < 1.25$} & \raisebox{-0.27ex}{$\delta < 1.25^2$} & \raisebox{-0.27ex}{$\delta < 1.25^3$}  \\
		\hline
            Frame(1) & 0.179 & 1.742 & 7.123 & 0.744 & 0.918 & 0.969 \\
		Frame(2) & 0.170 & 1.534 & 6.681 & 0.770  & 0.925 & 0.975 \\
            \textbf{Frame(3)} & \textbf{0.164} & \textbf{1.442} & \textbf{6.315} & \textbf{0.777} & \textbf{0.930} & \textbf{0.981}\\ 
		Frame(5) & 0.168 & 1.561 & 6.631  & 0.769 & 0.926  & 0.973  \\
		\hline
	\end{tabular}
        \vspace{-5.8mm}	
	\label{table:2}
\end{table}

\textbf{Impact of the Number of Spatiotemporal Priors Frames.} 
As shown in Table~\ref{table:2}, we can find that using one or two frames leads to poor performance since too less frames could not provide sufficient temporal features. While using more frames will also lead to a performance drop, since the accumulated inter-frame errors will increase the day-night differences. Using three frames can provide a coherent temporal structure that balances semantic context and temporal consistency, achieving the best performance.

\section{CONCLUSIONS AND FUTURE WORK}
In this paper, we present DASP, a self-supervised framework that exploits spatiotemporal priors for nighttime monocular depth estimation. Specifically, we first develop an adversarial network where the discriminator consists of four spatiotemporal priors learning blocks (SPLB). Particularly, the SPLB includes a spatial-based temporal learning module (STLM) to capture the motion-related variations along the time axis, and an axial spatial learning module (ASLM) to extract the spatial depth representation. The combination of STLM and ASLM provides sufficient spatiotemporal features for depth estimation. And then we devise a 3D projection consistency loss to strengthen geometric consistency and daytime priors. Extensive experiments conducted on two mainstream datasets demonstrate the effectiveness and stability of our method for nighttime depth estimation. In the future, we will further investigate the robustness of our model, especially in intense lighting environments and heavily blurred scenes.

\bibliographystyle{IEEEtran}
\bibliography{IEEEabrv.bib, IEEEexample.bib}

\end{document}